\pdfoutput=1
\RequirePackage[hyphens]{url}
\documentclass[11pt]{article}

\usepackage[]{emnlp2021}

\usepackage{times}
\usepackage{latexsym}

\usepackage[T1]{fontenc}

\usepackage[utf8]{inputenc}

\usepackage{microtype}

\usepackage{amssymb}
\usepackage{enumitem}

\newcommand{\mg}[1]{\dtcolornote[MG]{purple}{#1}}

\newcommand{\jb}[1]{\dtcolornote[JB]{brown}{#1}}

\newcommand{\bb}{\mathbf{b}}
\newcommand{\rr}{\mathbf{r}}
\newcommand{\xx}{\mathbf{x}}
\newcommand{\yy}{\mathbf{y}}
\newcommand{\hh}{\mathbf{h}}
\newcommand{\mm}{\mathbf{m}}
\newcommand{\oo}{\mathbf{o}}
\newcommand{\kk}{\mathbf{k}}
\newcommand{\vv}{\mathbf{v}}
\newcommand{\pp}{\mathbf{p}}

\usepackage{amsmath}
\usepackage{multirow}
\usepackage{booktabs}
\usepackage{graphicx}
\usepackage[normalem]{ulem}
\usepackage[cal=boondoxo]{mathalfa}

\DeclareMathOperator*{\argmax}{argmax}

\usepackage{amsthm}

\theoremstyle{definition}

\newif\ifcomments
\commentstrue
\ifcomments
    \providecommand{\mg}[1]{{\small \color{teal} [MG: #1]}}
    \providecommand{\rs}[1]{{\small \color{red} [RS: #1]}}
    \providecommand{\ol}[1]{{\small \color{orange} [OL: #1]}}
    \providecommand{\jb}[1]{{\small \color{blue} [JB: #1]}}
\else
    \providecommand{\mg}[1]{}
    \providecommand{\rs}[1]{}
    \providecommand{\ol}[1]{}
    \providecommand{\jb}[1]{}  
    \fi
\newcommand\nl[1]{{\it``#1''}}

\newcommand{\bluetarget}[1]{{\color{blue} #1}}

\title{Transformer Feed-Forward Layers Are Key-Value Memories}

\author{Mor Geva$^{1,2}$ ~~~~~ Roei Schuster$^{1,3}$ ~~~~~ Jonathan Berant$^{1,2}$ ~~~~~
Omer Levy$^{1}$ \\
$^1$Blavatnik School of Computer Science, Tel-Aviv University \\
$^2$Allen Institute for Artificial Intelligence \\
$^3$Cornell Tech \\
\small{\texttt{\{morgeva@mail,joberant@cs,levyomer@cs\}.tau.ac.il}},  \small{\texttt{rs864@cornell.edu}}
}

\begin{document}
\maketitle

\begin{abstract}
Feed-forward layers constitute two-thirds of a transformer model's parameters, yet their role in the network remains under-explored. 
We show that feed-forward layers in transformer-based language models operate as key-value memories, where each key correlates with textual patterns in the training examples, and each value induces a distribution over the output vocabulary. 
Our experiments show that the learned patterns are human-interpretable, and that lower layers tend to capture shallow patterns, while upper layers learn more semantic ones.
The values complement the keys' input patterns by inducing output distributions that concentrate probability mass on tokens likely to appear immediately after each pattern, particularly in the upper layers.
Finally, we demonstrate that the output of a  feed-forward layer is a composition of its memories, which is subsequently refined throughout the model's layers via residual connections to produce the final output distribution.
\end{abstract}

\begin{figure}[t]
\centering
\includegraphics[width=\columnwidth]{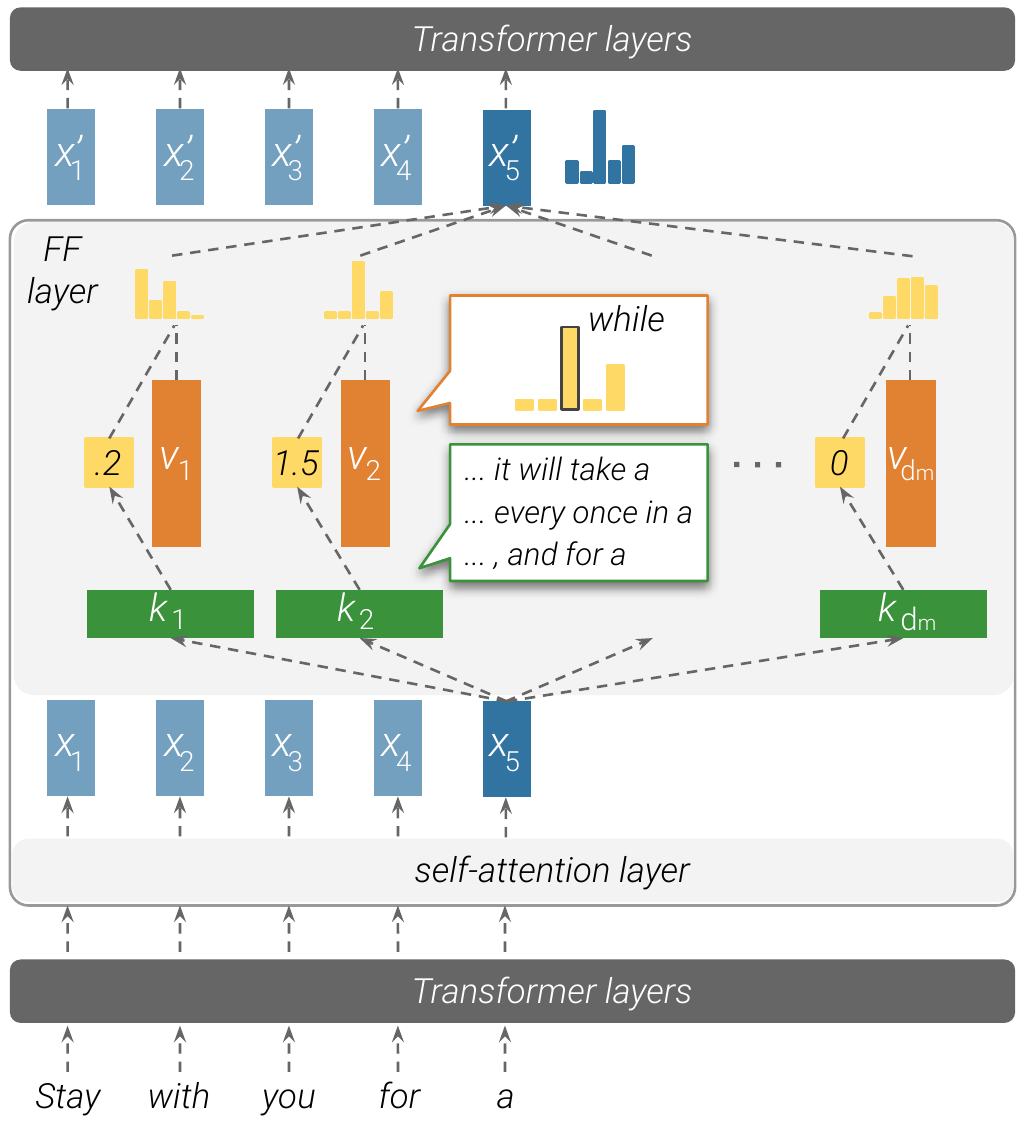}
\caption{An illustration of how a feed-forward layer emulates a key-value memory. Input vectors (here, $\xx_5$) are multiplied by \emph{keys} to produce \emph{memory coefficients} (e.g., the memory coefficient for $\vv_1$ is $0.2$), which then weigh distributions over the output vocabulary, stored in the \emph{values}. The feed-forward layer's output is thus the weighted sum of its values.
}
\label{figure:hypothesis}
\end{figure}

\section{Introduction}
\label{sec:introduction}

Transformer-based language models \cite{vaswani2017attention} are at the core of state-of-the-art natural language processing \cite{devlin2018bert, brown2020language}, largely due to the success of self-attention.
While much literature has been devoted to analyzing the function of self-attention layers \cite{voita2019analyzing, clark2019what, vig2019analyzing}, they account for only a third of a typical transformer's parameters ($4d^2$ per layer, where $d$ is the model's hidden dimension).
Most of the parameter budget is spent on position-wise feed-forward layers ($8d^2$ per layer), yet their role remains under-explored.
What, if so, is the function of feed-forward layers in a transformer language model?

We show that feed-forward layers emulate neural memories \cite{sukhbaatar2015end}, where the
first parameter matrix in the layer corresponds to \textit{keys}, and the second parameter matrix to \textit{values}.
Figure~\ref{figure:hypothesis} shows how the keys (first parameter matrix) interact with the input to produce coefficients, which are then used to compute a weighted sum of the values (second parameter matrix) as the output.
While the theoretical similarity between feed-forward layers and key-value memories has previously been suggested by \citet{sukhbaatar2019}, we take this observation one step further, and analyze the ``memories'' that the feed-forward layers store.

We find that each key correlates with a specific set of human-interpretable input patterns, such as $n$-grams or semantic topics. 
For example, $k_2$ in Figure~\ref{figure:hypothesis} is triggered by inputs that describe a period of time and end with \nl{a}.
Simultaneously, we observe that each \textit{value} can induce a distribution over the output vocabulary, and that this distribution correlates with the next-token distribution of the corresponding keys in the upper layers of the model.
In the above example, the corresponding value $v_2$ represents a distribution that puts most of its probability mass on the word \nl{while}.

Lastly, we analyze how the language model, as a whole, composes its final prediction from individual memories.
We observe that each layer combines hundreds of active memories, creating a distribution that is qualitatively different from each of its component memories' values.
Meanwhile, the residual connection between layers acts as a refinement mechanism, gently tuning the prediction at each layer while retaining most of the residual's information.


In conclusion, our work sheds light on the function of feed-forward layers in transformer-based language models. We show that feed-forward layers act as pattern detectors over the input across all layers, and that the final output distribution is gradually constructed in a bottom-up fashion.\footnote{The code for reproducing our experiments is available at \url{https://github.com/mega002/ff-layers/}.}







\section{Feed-Forward Layers as Unnormalized Key-Value Memories}
\label{sec:hypothesis}

\paragraph{Feed-forward layers}
A transformer language model \cite{vaswani2017attention} is made of intertwined self-attention and feed-forward layers.
Each feed-forward layer is a position-wise function, processing each input vector independently.
Let $\xx \in \mathbb{R}^d$ be a vector corresponding to some input text prefix.
We can express the feed-forward layer $\text{FF}(\cdot)$ as follows (bias terms are omitted):
\begin{align}
\text{FF}(\xx) = f(\xx \cdot K^{\top}) \cdot V
\label{eq:ffn}
\end{align}
Here, $K, V \in \mathbb{R}^{d_m \times d}$ are parameter matrices, and $f$ is a non-linearity such as ReLU.

\paragraph{Neural memory}
A neural memory \cite{sukhbaatar2015end} consists of $d_m$ key-value pairs, which we call \textit{memories}.\footnote{We use the terms ``memory cells'' and ``memories'' interchangeably.}
Each key is represented by a $d$-dimensional vector $\kk_i \in \mathbb{R}^d$, and together form the parameter matrix $K \in \mathbb{R}^{d_m \times d}$; likewise, we define the value parameters as $V \in \mathbb{R}^{d_m \times d}$.
Given an input vector $\xx \in \mathbb{R}^d$, we compute a distribution over the keys, and use it to compute the expected value:
\begin{align*}
p(k_i \mid x) &\propto \exp(\xx \cdot \kk_i) \\
\text{MN}(\xx) &= \sum_{i=1}^{d_m} p(k_i \mid x) \vv_i
\end{align*}
With matrix notation, we arrive at a more compact formulation:
\begin{align}
\text{MN}(\xx) &= \text{softmax}(\xx \cdot K^{\top}) \cdot V
\label{eq:mn}
\end{align}

\paragraph{Feed-forward layers emulate neural memory} Comparing equations \ref{eq:ffn} and \ref{eq:mn} shows that feed-forward layers are almost identical to key-value neural memories;
the only difference is that neural memory uses softmax as the non-linearity $f(\cdot)$, while the canonical transformer does not use a normalizing function in the feed-forward layer.
The \textit{hidden dimension} $d_m$ is essentially the number of memories in the layer, and the activation $\mm = f(\xx \cdot K^\top)$, commonly referred to as the \textit{hidden layer}, is a vector containing an unnormalized non-negative coefficient for each memory.
We refer to each $\mm_i$ as the \textit{memory coefficient} of the $i$th memory cell.

\citet{sukhbaatar2019} make an analogous observation, and incorporate the parameters of the feed-forward layers as persistent memory cells in the self-attention layers.
While this reparameterization works in practice, the experiment does not tell us much about the role of feed-forward layers in the canonical transformer.
If transformer feed-forward layers are indeed key-value memories, then what memories do they store?

We conjecture that each key vector $\kk_i$ captures a particular pattern (or set of patterns) in the input sequence (Section~\ref{sec:w1_as_keys}), and that its corresponding value vector $\vv_i$ represents the distribution of tokens that follows said pattern (Section~\ref{sec:w2_as_values}).


\section{Keys Capture Input Patterns}
\label{sec:w1_as_keys}

\begin{table*}[t]
    \centering
    \footnotesize
    \begin{tabular}{l|p{2.8cm}|p{10.cm}}
         Key & Pattern & Example trigger prefixes  \\ \toprule
         \multirow{3}{*}{$\kk_{449}^{1}$} & \multirow{3}{=}{Ends with \nl{substitutes} (\textcolor{blue}{shallow})} &  \emph{At the meeting, Elton said that ``for artistic reasons there could be no substitutes} \\ 
         & & \emph{In German service, they were used as substitutes} \\ 
         & & \emph{Two weeks later, he came off the substitutes} \\ \hline 
        \multirow{3}{*}{$\kk_{2546}^{6}$} & \multirow{3}{=}{Military, ends with \nl{base}/\nl{bases} (\textcolor{blue}{shallow + semantic})} & \emph{On 1 April the SRSG authorised the SADF to leave their bases}\\
         & & \emph{Aircraft from all four carriers attacked the Australian base} \\
         & & \emph{Bombers flying missions to Rabaul and other Japanese bases} \\ \hline
         \multirow{3}{*}{$\kk_{2997}^{10}$} & \multirow{3}{=}{a ``part of'' relation (\textcolor{blue}{semantic})} & \emph{In June 2012 she was named as one of the team that competed} \\
         & & \emph{He was also a part of the Indian delegation} \\
         & & \emph{Toy Story is also among the top ten in the BFI list of the 50 films you should} \\ \hline
         \multirow{3}{*}{$\kk_{2989}^{13}$} & \multirow{3}{=}{Ends with a time range (\textcolor{blue}{semantic})} & \emph{Worldwide, most tornadoes occur in the late afternoon, between 3 pm and 7} \\
         & & \emph{Weekend tolls are in effect from 7:00 pm Friday until} \\
         & & \emph{The building is open to the public seven days a week, from 11:00 am to} \\ \hline
         \multirow{3}{*}{$\kk_{1935}^{16}$} & \multirow{3}{=}{TV shows (\textcolor{blue}{semantic})} & \emph{Time shifting viewing added 57 percent to the episode's} \\
         & & \emph{The first season set that the episode was included in was as part of the} \\
         & & \emph{From the original NBC daytime version , archived} \\ \hline
    \end{tabular}
    \caption{Examples of human-identified patterns that trigger different memory keys.}
    \label{table:pattern_examples}
\end{table*}

We posit that the key vectors $K$ in feed-forward layers act as pattern detectors over the input sequence, where each individual key vector $\kk_i$ corresponds to a specific pattern over the input prefix $x_1, \ldots, x_j$.
To test our claim, we analyze the keys of a trained language model's feed-forward layers.
We first retrieve the training examples (prefixes of a sentence)
most associated with a given key, that is, the input texts where the memory coefficient is highest. 
We then ask humans to identify patterns within the retrieved examples.
For almost 
every key $\kk_i$ in our sample, a small set of well-defined patterns, recognizable by humans, covers most of the examples associated with the key.

\subsection{Experiment}

We conduct our experiment over the language model of \citet{baevski2018adaptive}, a 16-layer transformer language model trained on WikiText-103 \cite{merity2017pointer}.
This model defines $d=1024$ and $d_m=4096$, and has a total of $d_m\cdot 16=65,536$ potential keys to analyze.
We randomly sample 10 keys per layer (160 in total).

\paragraph{Retrieving trigger examples}
We assume that patterns stored in memory cells originate from examples the model was trained on. Therefore, given a key $\kk_i^\ell$ that corresponds to the $i$-th hidden dimension of the $\ell$-th feed-forward layer, we compute the memory coefficient $\text{ReLU} (\xx_j^\ell \cdot \kk_i^\ell)$ for every prefix $x_1,\ldots,x_j$ of every sentence from the WikiText-103's training set.\footnote{We segment training examples into sentences to simplify the annotation task and later analyses.} For example, for the hypothetical sentence \nl{I love dogs}, we will compute three coefficients, for the prefixes \nl{I}, \nl{I love}, and \nl{I love dogs}. Then, we retrieve the \emph{top-$t$ trigger examples}, that is, the $t$ prefixes whose representation at layer $\ell$ yielded the highest inner product with $\kk_i^\ell$.

\begin{figure}[t]
    \centering
    \includegraphics[width=\columnwidth]{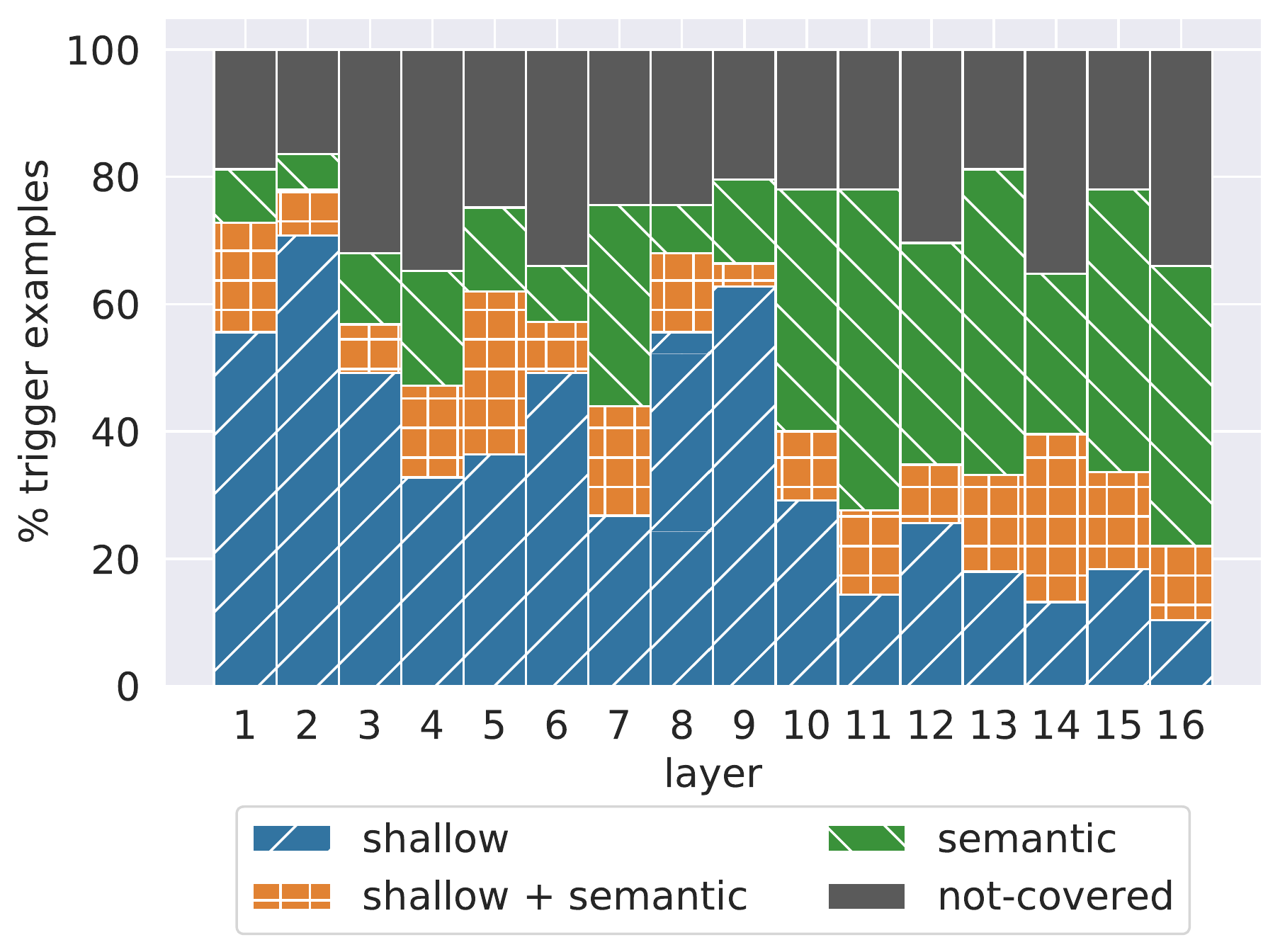}
    \caption{Breakdown of the labels experts assigned to trigger examples in each layer. Some examples were not associated with any pattern (``not-covered'').
    }
    \label{figure:pattern_annotations}
\end{figure}

\begin{figure}[t]
    \centering
    \includegraphics[scale=0.4]{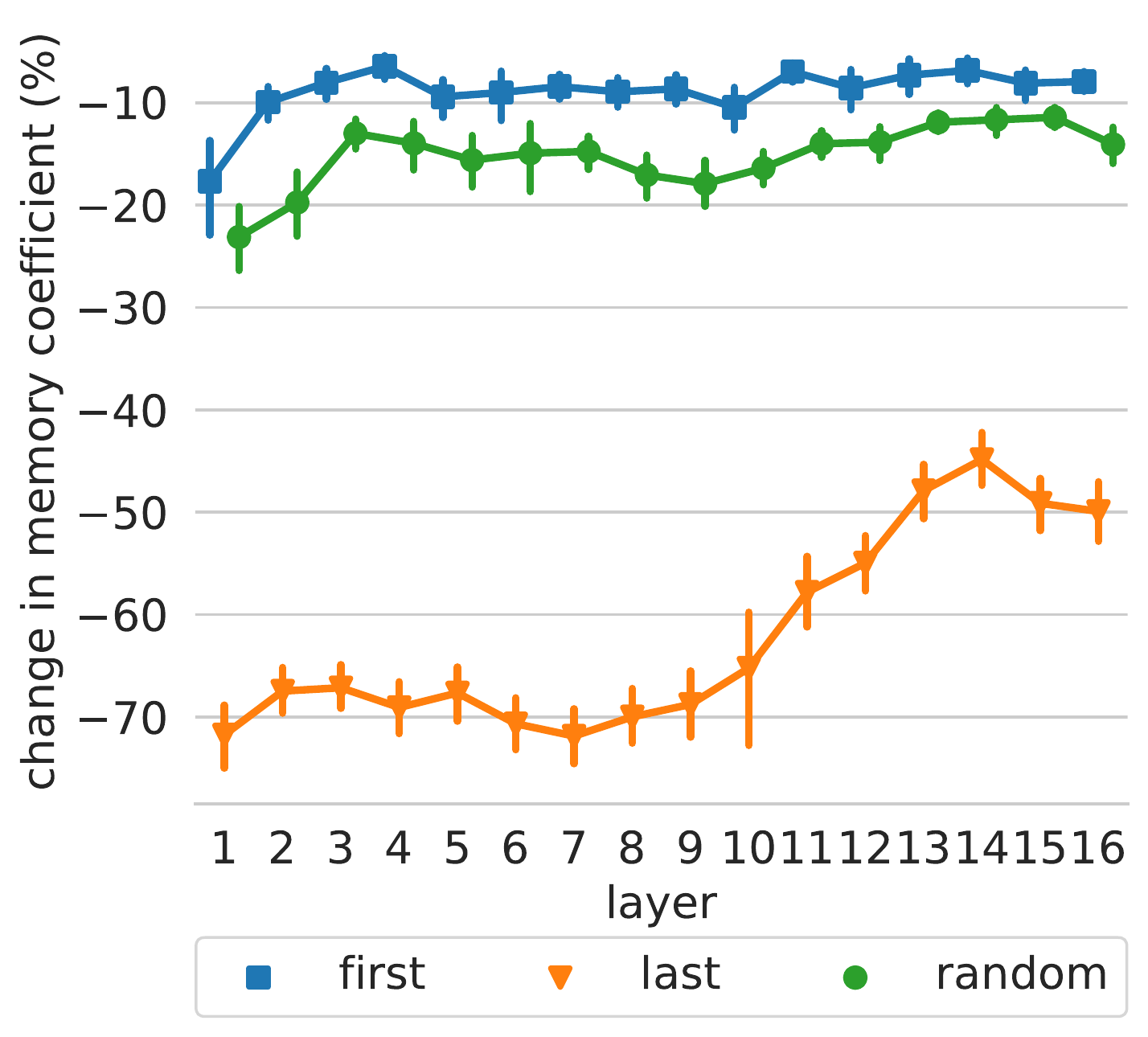}
    \caption{Relative change in memory coefficient caused by removing the first, the last, or a random token from the input.}
    \label{figure:input_modifications}
\end{figure}

\paragraph{Pattern analysis} 
We let human experts (NLP graduate students) annotate the top-25 prefixes retrieved for each key, and asked them to (a) identify repetitive patterns that occur in at least 3 prefixes (which would strongly indicate a connection to the key, as this would unlikely happen if sentences were drawn at random) (b) describe each recognized pattern, and (c) classify each recognized pattern as \nl{shallow} (e.g. recurring n-grams) or \nl{semantic} (recurring topic).
Each key and its corresponding top-25 prefixes were annotated by one expert.
To assure that every pattern is grounded in at least 3 prefixes, we instruct the experts to specify, for each of the top-25 prefixes, which pattern(s) it contains. A prefix may be associated with multiple (shallow or semantic) patterns. 

Table~\ref{table:pattern_examples} shows example patterns.
A fully-annotated example of the top-25 prefixes from a single memory key 
is shown in Appendix~\ref{subsec:pattern_analysis_full_example}.

\subsection{Results}

\paragraph{Memories are associated with human-recognizable patterns}
Experts were able to identify at least one pattern for every key, with an average of 3.6 identified patterns per key.
Furthermore, the vast majority of retrieved prefixes (65\%-80\%) were associated with at least one identified pattern (Figure~\ref{figure:pattern_annotations}).
Thus, the top examples triggering each key share clear patterns that humans can recognize.


\paragraph{Shallow layers detect shallow patterns}
Comparing the amount of prefixes associated with shallow patterns and semantic patterns (Figure~\ref{figure:pattern_annotations}), the lower layers (layers 1-9) are dominated by shallow patterns, often with prefixes that share the last word (e.g. $\kk_{449}^{1}$ in Table~\ref{table:pattern_examples}). In contrast, the upper layers (layers 10-16) are characterized by more semantic patterns, with prefixes from similar contexts but without clear surface-form similarities (e.g. $\kk_{1935}^{16}$ in Table~\ref{table:pattern_examples}).
This observation corroborates recent findings that lower (upper) layers in deep contextualized models encode shallow (semantic) features of the inputs
\cite{peters2018elmo,jawahar2019bert, liu2019linguistic}.

To further test this hypothesis, we sample 1600 random keys (100 keys per layer) and apply local modifications to the top-50 trigger examples of every key.
Specifically, we remove either the \textit{first}, \textit{last}, or a \textit{random} token from the input, and measure how this mutation affects the memory coefficient.
Figure~\ref{figure:input_modifications} shows that the model considers the end of an example as more salient than the beginning for predicting the next token. 
In upper layers, removing the last token has less impact, supporting our conclusion that upper-layer keys are less correlated with shallow patterns.



\section{Values Represent Distributions}
\label{sec:w2_as_values}

\begin{figure}[t]
    \centering
    \includegraphics[width=\columnwidth]{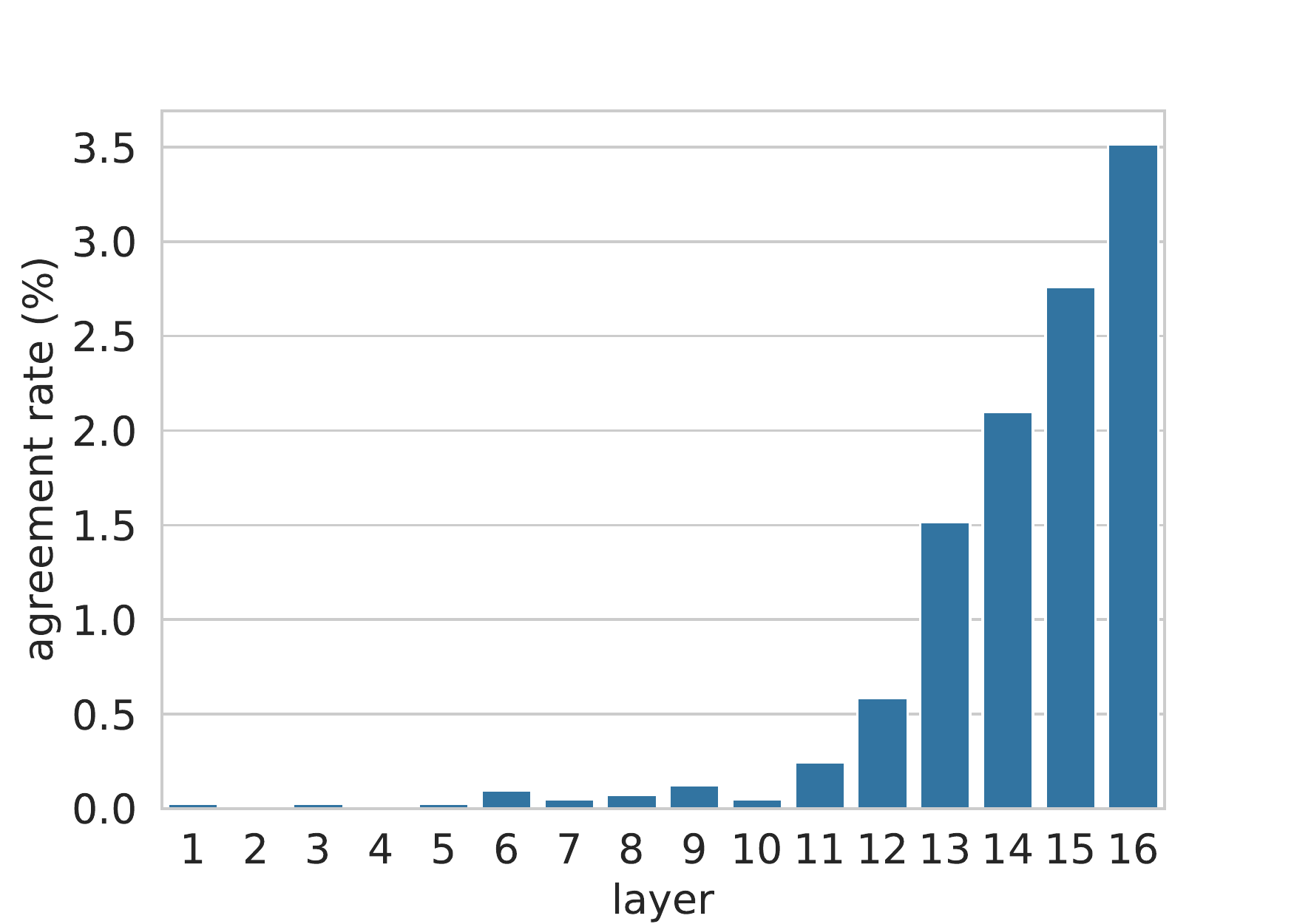}
    \caption{Agreement rate between the top-ranked token based on the value vector $\vv_i^\ell$, and the next token of the top-ranked trigger example associated with the key vector $\kk_i^\ell$.}
    \label{figure:w2_embedding_match_top_target} 
    
\end{figure}

After establishing that keys capture patterns in training examples, we turn to analyze the information stored in their corresponding values.
We show that each value $\vv_i^\ell$ can be viewed as a distribution over the output vocabulary, and demonstrate that this distribution complements the patterns in the corresponding key $\kk_i^\ell$ in the model's upper layers (see Figure~\ref{figure:hypothesis}). 

\paragraph{Casting values as distributions over the vocabulary.}
We begin by converting each value vector $\vv_i^\ell$ into a probability distribution over the vocabulary by multiplying it by the output embedding matrix $E$ and applying a softmax:\footnote{This is a simplification; in practice, we use the adaptive softmax \cite{baevski2018adaptive} to compute probabilities.}
\begin{align*}
\pp_i^\ell = \text{softmax} (\vv_i^\ell \cdot E).
\end{align*}
The probability distribution $\pp_i^\ell$ is uncalibrated, since the value vector $\vv_i^\ell$ is typically multiplied by the input-dependent memory coefficient $\mm_i^\ell$, changing the skewness of the output distribution.
That said, the \textit{ranking} induced by $\pp_i^\ell$ is invariant to the coefficient, and can still be examined.
This conversion assumes (na\"ively) that all model's layers operate in the same embedding space.

\begin{figure}[t]
    \centering
    \includegraphics[width=\columnwidth]{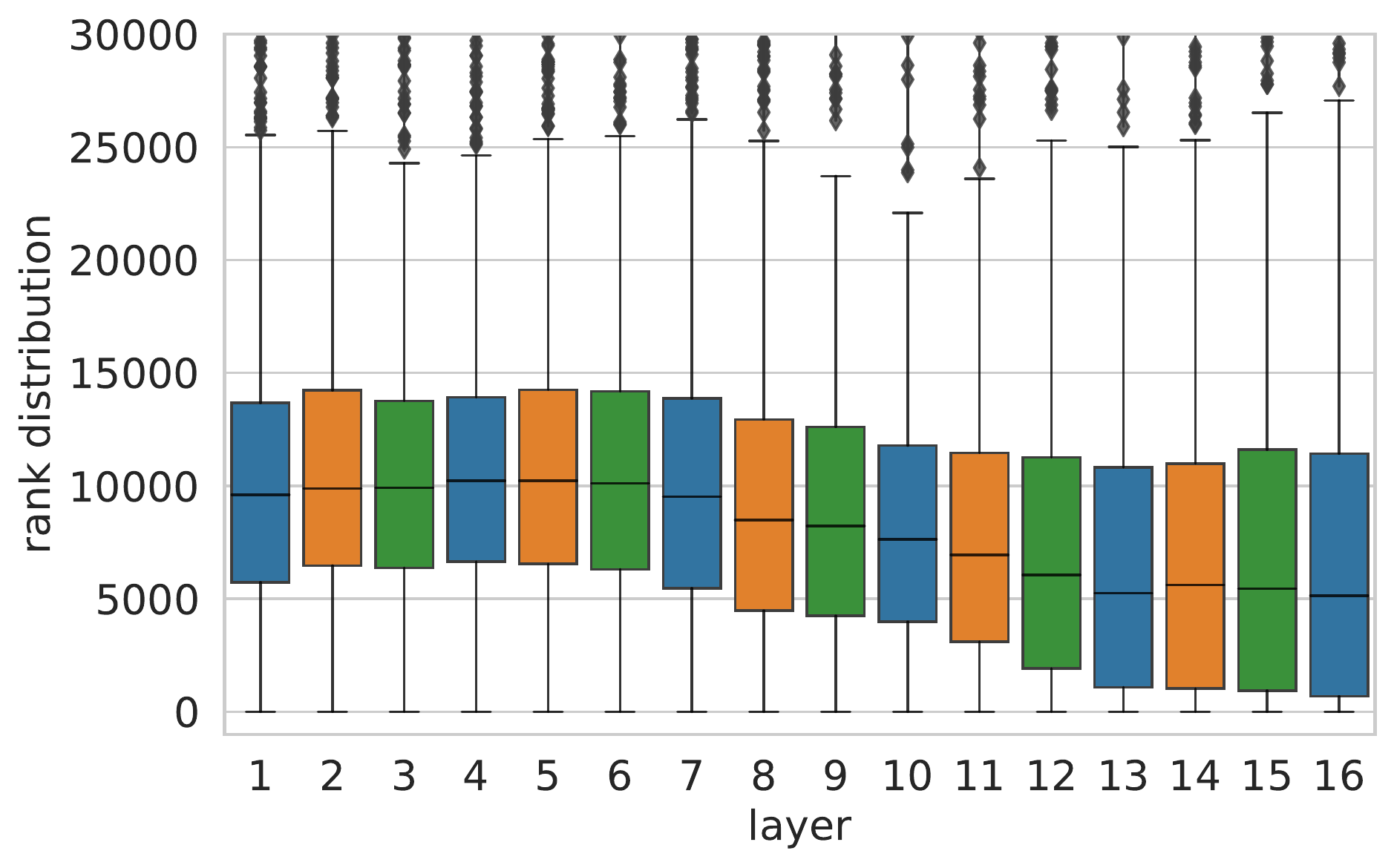}
    \caption{Distribution of the rank of the next-token in the top-1 trigger example of $\kk_i^\ell$ ($w_i^\ell$), according to the ranking induced by the value vector $\vv_i^\ell$. We cut the tail of the distribution, which stretches up to the vocabulary size ($\sim$270K tokens).
    }
    \label{figure:target_rank_in_w2_embedding_dist}
    
\end{figure}

\paragraph{Value predictions follow key patterns in upper layers.}
For every layer $\ell$ and memory dimension $i$, we compare the top-ranked token according to $\vv_i^\ell$, ($\argmax(\pp_i^\ell)$) to the next token $w_i^\ell$ in the top-1 trigger example according to $\kk_i^\ell$ (the example whose memory coefficient for $\kk_i^\ell$ is the highest).
Figure~\ref{figure:w2_embedding_match_top_target} shows the \emph{agreement rate}, i.e. the fraction of memory cells (dimensions) where the value's top prediction matches the key's top trigger example ($\argmax(\pp_i^\ell) = w_i^\ell$). 
It can be seen that the agreement rate is close to zero in the lower layers (1-10), but starting from layer 11, the agreement rate quickly rises until 3.5\%, showing higher agreement between keys and values on the identity of the top-ranked token.
Importantly, this value is orders of magnitude higher than a random token prediction from the vocabulary, which would produce a far lower agreement rate ($0.0004\%$), showing that upper-layer memories manifest non-trivial predictive power.

Next, we take the next token of $\kk_i^\ell$'s top-1 trigger example ($w_i^\ell$), and find where it ranks in the value vector's distribution $\pp_i^\ell$.
Figure~\ref{figure:target_rank_in_w2_embedding_dist} shows that the rank of the next token of a trigger example increases through the layers, meaning that $w_i^\ell$ tends to get higher probability in the upper layers.

\begin{figure}[t]
    \centering
    \includegraphics[scale=0.44]{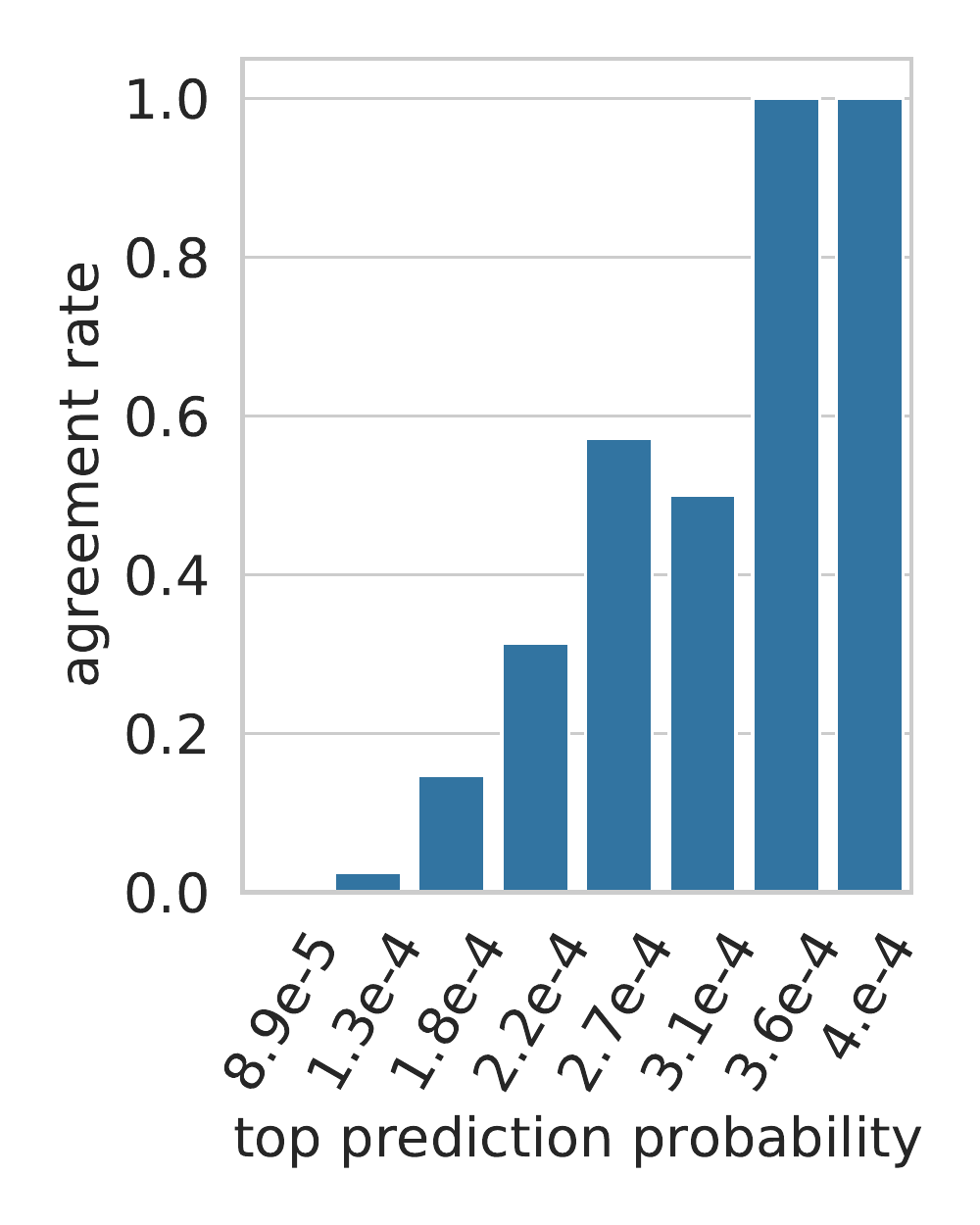}
    \caption{Agreement rate (between the top-ranked token based on the value vector $\vv_i^\ell$ and the next token of the top-ranked trigger example associated with the key vector $\kk_i^\ell$) as a function of the maximal probability assigned by the value vector.}
    \label{figure:agreement_rate_prediction_confidence} 
    
\end{figure}

\paragraph{Detecting predictive values.}
To examine if we can automatically detect values with high agreement rate, we analyze the probability of the values' top prediction, i.e., ($\max (\pp_i^\ell)$).
Figure~\ref{figure:agreement_rate_prediction_confidence} shows that although these distributions are not calibrated, distributions with higher maximum probabilities are more likely to agree with their key's top trigger example.
We then 
take the 100 values with highest probability across all layers and dimensions (97 out of the 100 are in the upper layers, 11-16), and
for each value $\vv_i^\ell$ , analyze the top-50 trigger examples of $\kk_i^\ell$. 
We find that in almost half of the values (46 out of 100), there is at least one trigger example that agrees with the value's top prediction. 
Examples are provided in Table~\ref{table:w2_embedding_match_in_top_targets}.

\paragraph{Discussion.} When viewed as distributions over the output vocabulary, values in the upper layers tend to assign higher probability to the next-token of examples triggering the corresponding keys.
This suggests that memory cells often store information on how to directly predict the output (the distribution of the next word) from the input (patterns in the prefix).
Conversely, the lower layers do not exhibit such clear correlation between the keys' patterns and the corresponding values' distributions.
A possible explanation is that the lower layers do not operate in the same embedding space, and therefore, projecting values onto the vocabulary using the output embeddings does not produce distributions that follow the trigger examples. However, our results imply that some intermediate layers \emph{do} operate in the same or similar space to upper layers (exhibiting some agreement), which in itself is non-trivial.
We leave further exploration of this phenomenon to future work.

\begin{table*}[t]
    \centering
    \footnotesize
    \begin{tabular}{llrp{10cm}}
        Value & Prediction & Precision@50 & Trigger example\\
        \toprule
        $\vv_{222}^{15}$ & \textit{each} & 68\% & \emph{But when bees and wasps resemble \bluetarget{each}} \\ \midrule
        $\vv_{752}^{16}$ & \textit{played} & 16\% & \emph{Her first role was in Vijay Lalwani's psychological thriller Karthik Calling Karthik, where Padukone was cast as the supportive girlfriend of a depressed man (\bluetarget{played}} \\ \midrule
        $\vv_{2601}^{13}$ & \textit{extratropical} & ~~4\% & \emph{Most of the winter precipitation is the result of synoptic scale, low pressure weather systems (large scale storms such as \bluetarget{extratropical}} \\ \midrule
        $\vv_{881}^{15}$ & \textit{part} & 92\% & \emph{Comet served only briefly with the fleet, owing in large \bluetarget{part}} \\ \midrule
        $\vv_{2070}^{16}$ & \textit{line} & 84\% & \emph{Sailing from Lorient in October 1805 with one ship of the \bluetarget{line}} \\ \midrule
        $\vv_{3186}^{12}$ & \textit{jail} & ~~4\% & \emph{On May 11, 2011, four days after scoring 6 touchdowns for the Slaughter, Grady was sentenced to twenty days in \bluetarget{jail}} \\
        \bottomrule
    \end{tabular}
    \caption{Example values, their top prediction, the fraction of their key's top-50 trigger examples that agree with their prediction, and a matching trigger example (with the target token marked in blue).}
    \label{table:w2_embedding_match_in_top_targets}
\end{table*}

\section{Aggregating Memories}

So far, our discussion has been about the function of a single memory cell in feed-forward layers.
How does the information from \textit{multiple} cells in multiple \textit{layers} aggregate to form a model-wide prediction?
We show that every feed-forward layer combines multiple memories to produce a distribution that is qualitatively different from each of its component memories' value distributions (Section~\ref{subsec:ffn_pattern_aggregation}).
These layer-wise distributions are then combined via residual connections in a refinement process, where each feed-forward layer updates the residual's distribution to finally form the model's output (Section~\ref{subsec:ffn_residual}).

\subsection{Intra-Layer Memory Composition}
\label{subsec:ffn_pattern_aggregation}

The feed-forward layer's output can be defined as the sum of value vectors weighted by their memory coefficients, plus a bias term:
\begin{align*}
\yy^\ell = \sum_i \text{ReLU} (\xx^\ell \cdot \kk_i^\ell ) \cdot \vv_i^\ell + \bb^\ell.
\end{align*}
If each value vector $\vv_i^\ell$ contains information about the target token's distribution, how is this information aggregated into a single output distribution?
To find out, we analyze the behavior of 4,000 randomly-sampled prefixes from the validation set. Here, the validation set is used (rather than the training set used to find trigger examples) since we are trying to characterize the model's behavior at inference time, not find the examples it ``memorizes'' during training.

\begin{figure}[t]
    \centering
    \includegraphics[scale=0.45, trim= 0.8cm 0cm 0cm 0cm]{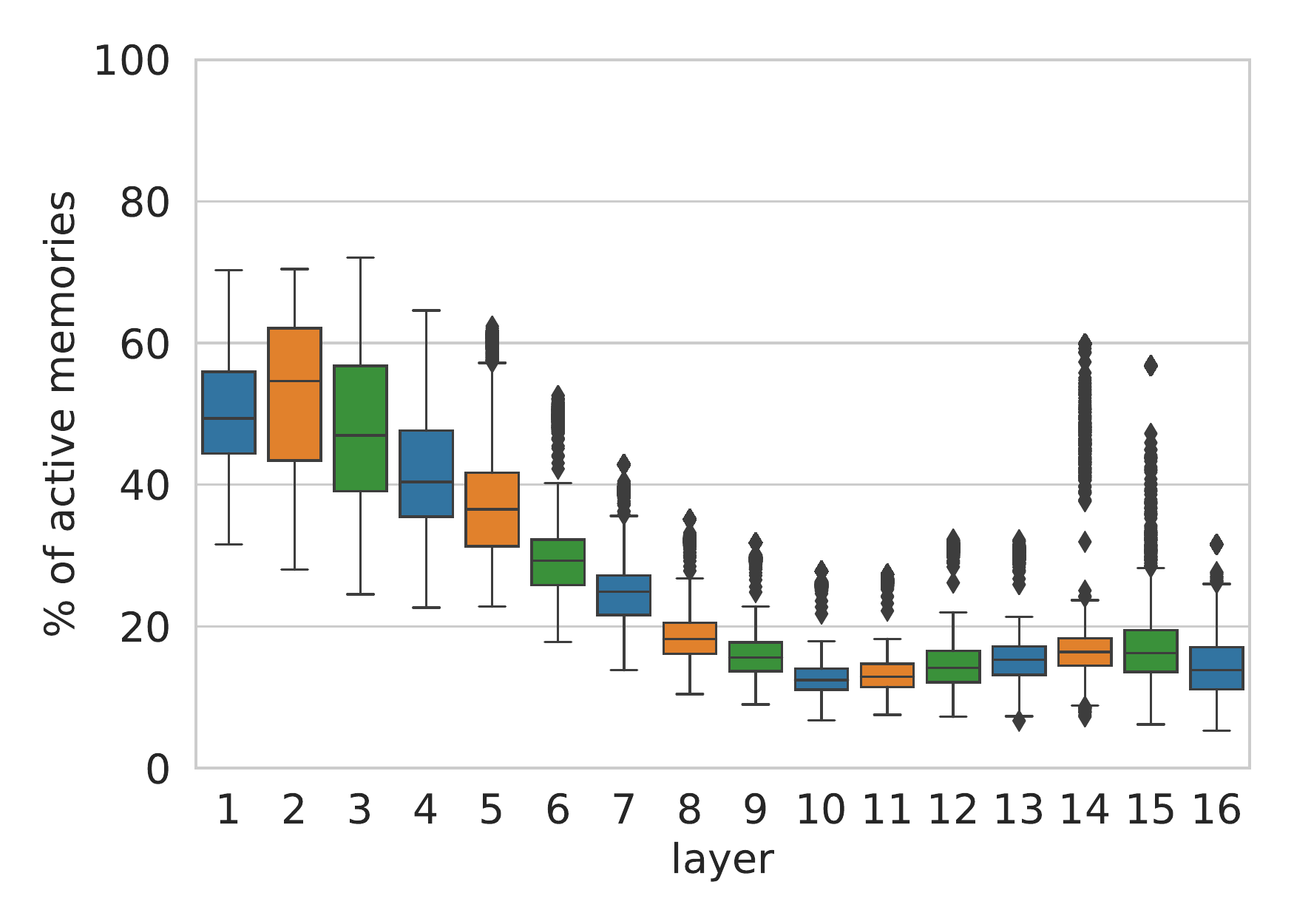}
    \caption{The fraction of active memories (i.e., with positive memory coefficient) out of 4096 memories in every layer, for a random sample of 4,000 examples.}
    \label{figure:coeffs_l0}
\end{figure}

We first measure the fraction of ``active'' memories (cells with a non-zero coefficient).
Figure~\ref{figure:coeffs_l0} shows that a typical example triggers hundreds of memories per layer (10\%-50\% of 4096 dimensions), but the majority of cells remain inactive.
Interestingly, the number of active memories drops towards layer 10, which is the same layer in which semantic patterns become more prevalent than shallow patterns, according to expert annotations (see Section~\ref{sec:w1_as_keys}, Figure~\ref{figure:pattern_annotations}).

\begin{figure}[t]
    \centering
    \includegraphics[width=\columnwidth]{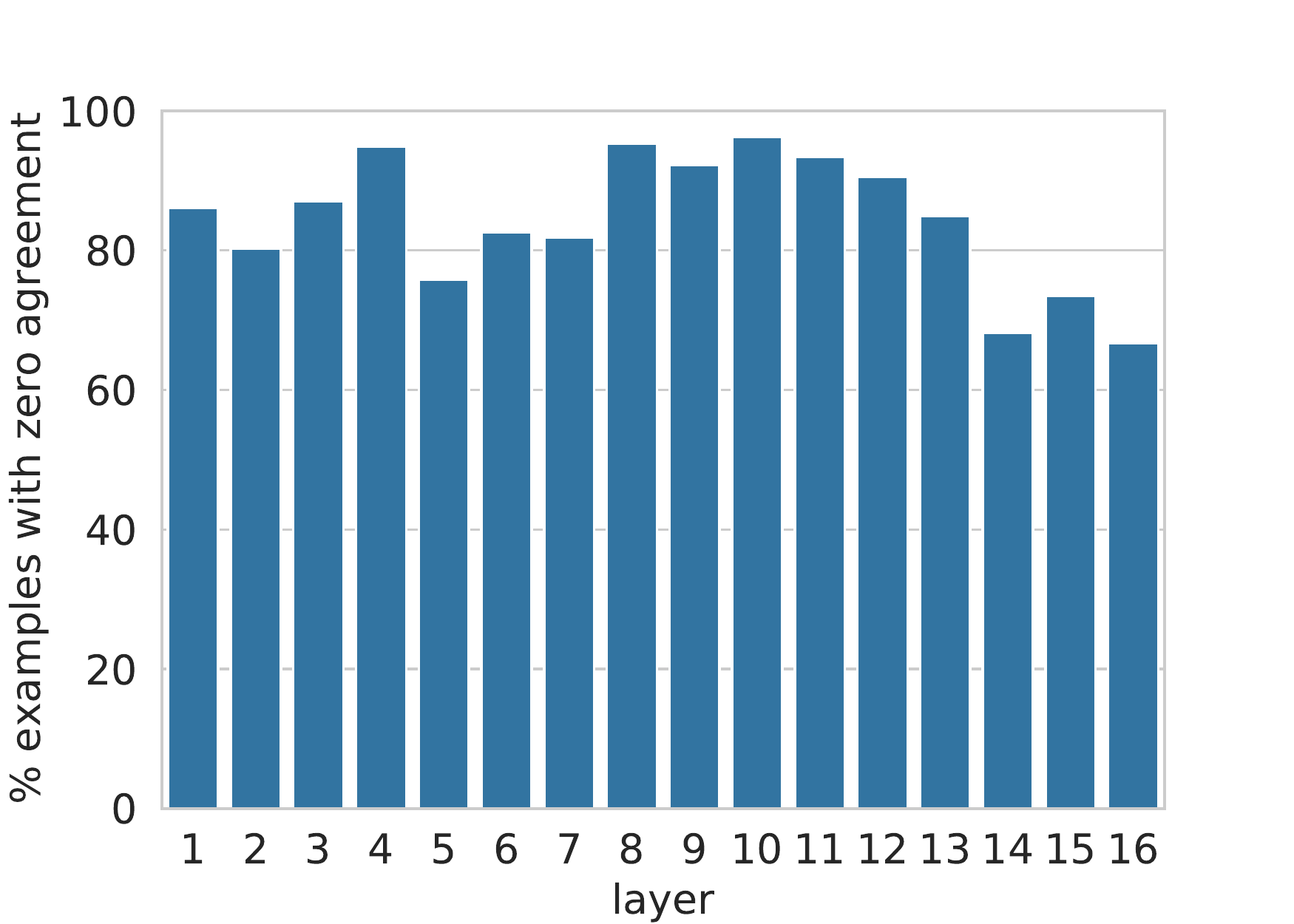}
    \caption{The fraction of examples in a random sample of 4,000 examples where the layer's prediction is different from the prediction of all of its memories. }
    \label{figure:compositionality}
\end{figure}

While there are cases where a single memory cell dominates the output of a layer, the majority of outputs are clearly compositional.
We count the number of instances where the feed-forward layer's top prediction is \textit{different} from all of the memories' top predictions. Formally, we denote:
\begin{align*}
\text{top}(\hh) = \argmax(\hh \cdot E)
\end{align*}
as a generic shorthand for the top prediction from the vocabulary distribution induced by the vector $\hh$, and compute the number of examples where the following condition holds:
\begin{align*}
\forall i : \text{top}(\vv^\ell_i) \neq \text{top}(\yy^\ell)
\end{align*}
Figure~\ref{figure:compositionality} shows that, for any layer in the network, the layer's final prediction is different than \emph{every one} of the memories' predictions in at least $\sim$68\% of the examples. Even in the upper layers, where the memories' values are more correlated with the output space (Section~\ref{sec:w2_as_values}), the layer-level prediction is typically not the result of a single dominant memory cell, but a composition of multiple memories.

We further analyze cases where at least one memory cell agrees with the layer's prediction, and find that (a) in 60\% of the examples the target token is a common stop word in the vocabulary (e.g. \nl{the} or \nl{of}), and (b) in 43\% of the cases the input prefix has less than 5 tokens. This suggests that very common patterns in the training data might be ``cached'' in individual memory cells, and do not require compositionality.

\subsection{Inter-Layer Prediction Refinement}
\label{subsec:ffn_residual}

While a single feed-forward layer composes its memories in parallel, a multi-layer model uses the residual connection $\rr$ to \textit{sequentially} compose predictions to produce the model's final output:\footnote{The residual propagates information from previous layers, including the transformer's self-attention layers.}
\begin{align*}
\xx^\ell &= \text{LayerNorm}(\rr^\ell) \\
\yy^\ell &= \text{FF}(\xx^\ell) \\
\oo^{\ell} &= \yy^\ell + \rr^\ell
\end{align*}
We hypothesize that the model uses the sequential composition apparatus as a means to \textit{refine} its prediction from layer to layer, often deciding what the prediction will be at one of the lower layers.

\begin{figure}[t]
    \centering
    \includegraphics[width=\columnwidth]{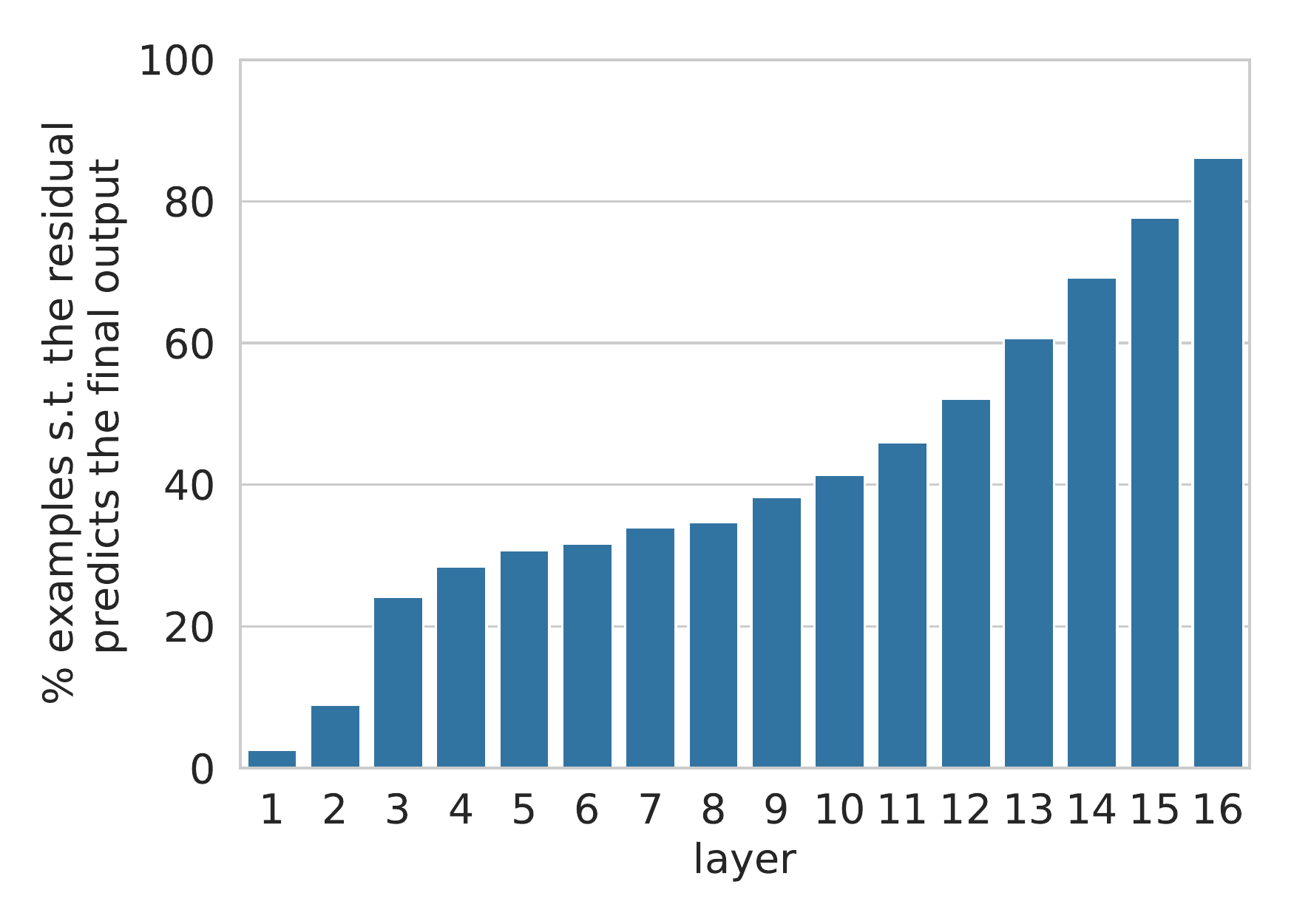}
    \caption{Fraction of examples in each layer, where the residual's top prediction matches the model's output.}
    \label{figure:residual_final_output_preds}
\end{figure}

To test our hypothesis, we first measure how often the probability distribution induced by the residual vector $\rr^\ell$ matches the model's final output $\oo^{L}$ ($L$ being the total number of layers):
\begin{align*}
\text{top}(\rr^\ell) = \text{top}(\oo^{L})
\end{align*}
Figure~\ref{figure:residual_final_output_preds} shows that roughly a third of the model's predictions are determined in the bottom few layers.
This number grows rapidly from layer 10 onwards, implying that the majority of ``hard'' decisions occur before the final layer.

\begin{figure}[t]
    \centering
    \includegraphics[width=\columnwidth]{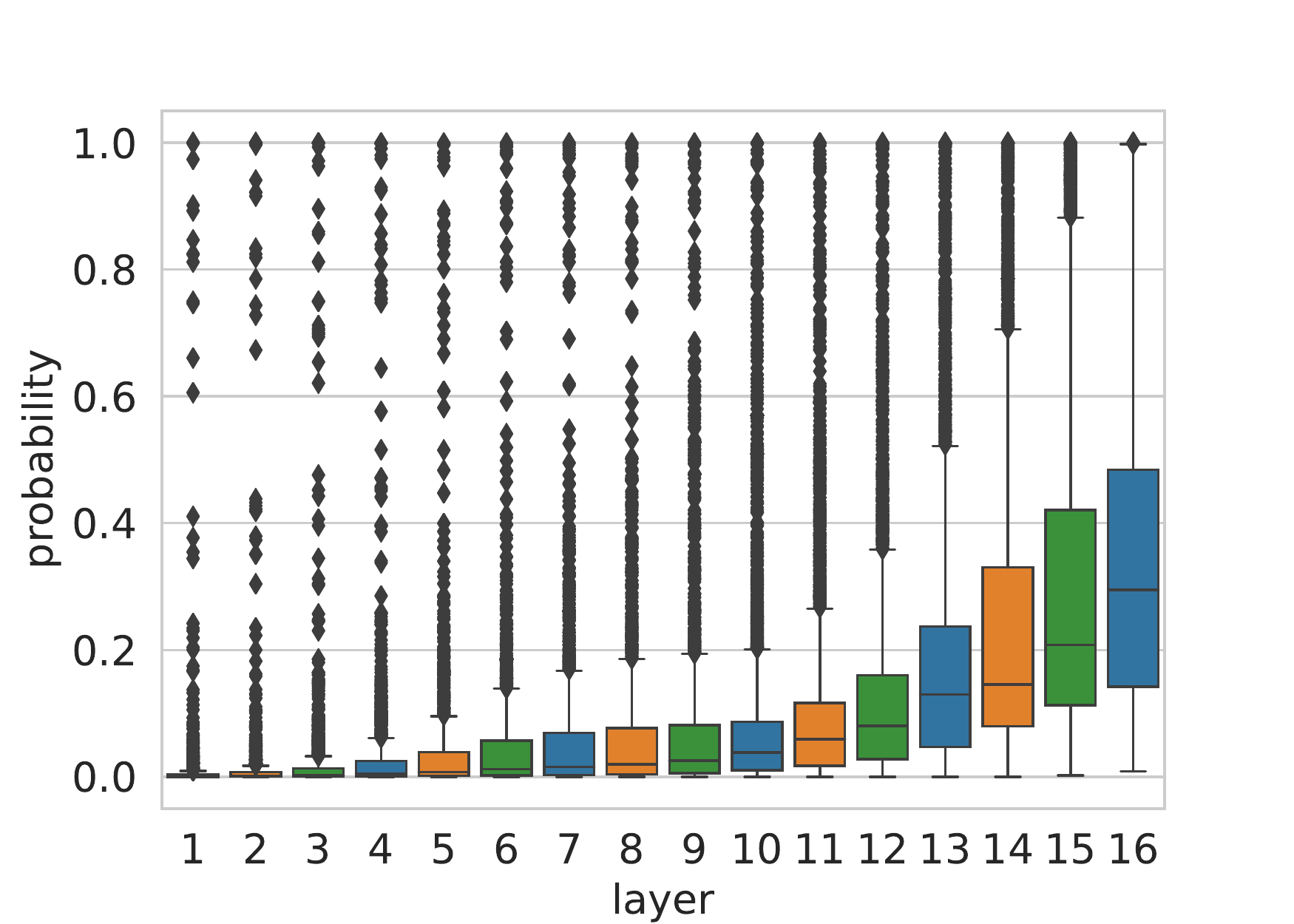}
    \caption{Probability of the token output by the model according to the residual of each layer.}
    \label{figure:residual_output_prob}
\end{figure}

We also measure the probability mass $p$ that each layer's residual vector $\rr^\ell$ assigns to the model's final prediction:
\begin{align*}
w &= \text{top}(\oo^{L}) \\
\pp &= \text{softmax} (\rr^\ell \cdot E) \\
p &= \pp_w
\end{align*}
Figure~\ref{figure:residual_output_prob} shows a similar trend, but emphasizes that it is not only the top prediction's identity that is refined as we progress through the layers, it is also the model's confidence in its decision.

To better understand how the refinement process works at each layer, we measure how often the residual's top prediction changes following its interaction with the feed-forward layer ($\text{top}(\rr^\ell) \neq \text{top}(\oo^{\ell})$), and whether this change results from the feed-forward layer overriding the residual ($\text{top}(\oo^{\ell}) = \text{top}(\yy^\ell)$) or from a true composition ($\text{top}(\rr^\ell) \neq \text{top}(\oo^{\ell}) \neq \text{top}(\yy^\ell)$).

\begin{figure}[t]
    \centering
    \includegraphics[width=\columnwidth]{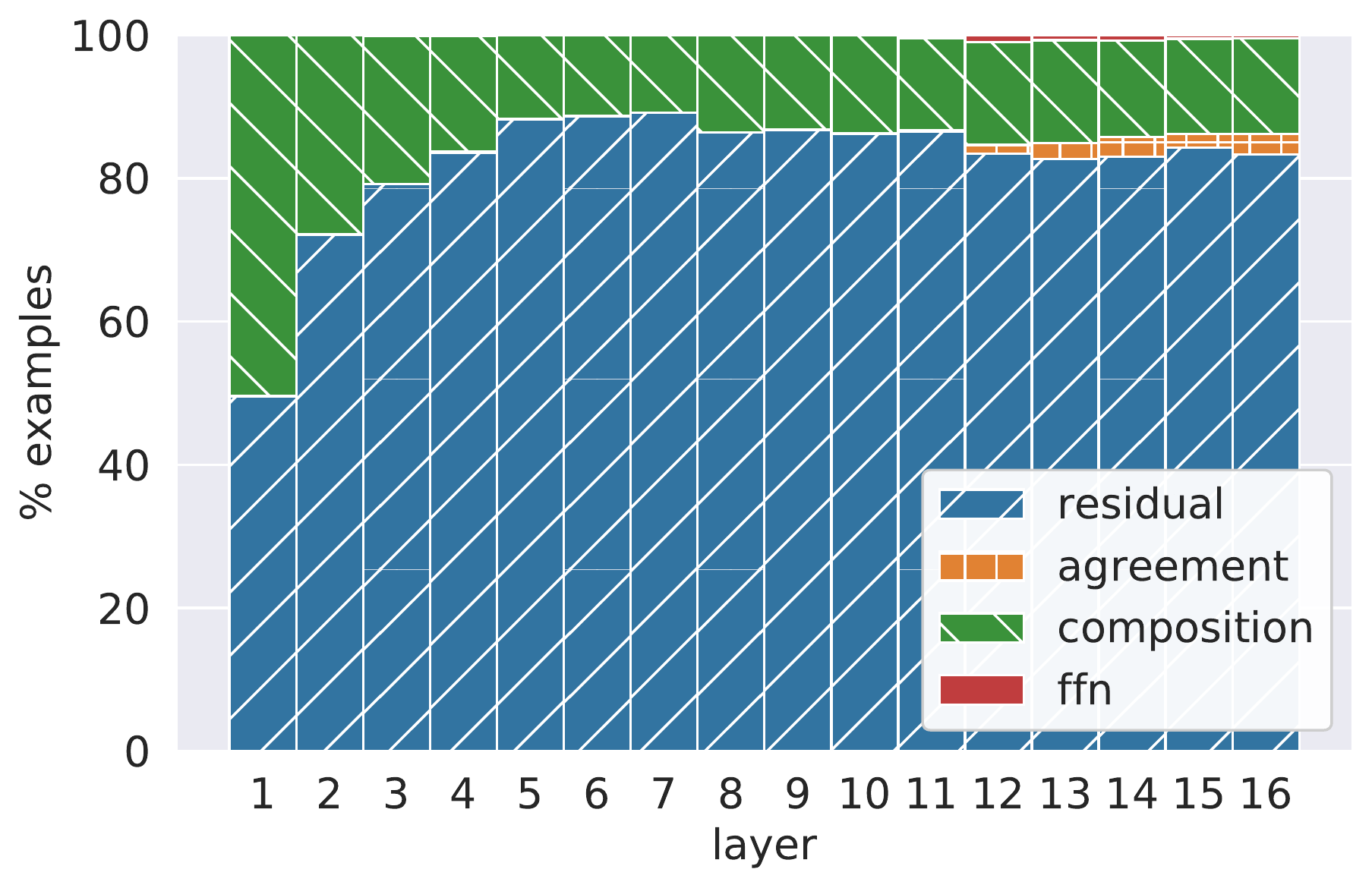}
    \caption{Breakdown of examples by prediction cases: the layer's output prediction matches the residual's prediction (\emph{residual}), matches the feed-forward layer's prediction (\emph{ffn}), matches both of the predictions (\emph{agreement}), or none of the predictions (\emph{composition}). By construction, there are no cases where the residual's prediction matches the feed-forward layer's prediction and does not match the output's prediction.}
    \label{figure:prediction_cases}
\end{figure}

Figure~\ref{figure:prediction_cases} shows the breakdown of different cases per layer.
In the vast majority of examples, the residual's top prediction ends up being the model's prediction (\textit{residual+agreement}). In most of these cases, the feed forward layer predicts something different (\textit{residual}).
Perhaps surprisingly, when the residual's prediction does change (\textit{composition+ffn}), it rarely changes to the feed-forward layer's prediction (\textit{ffn}).
Instead, we observe that composing the residual's distribution with that of the feed-forward layer produces a ``compromise'' prediction, which is equal to neither (\textit{composition}).
This behavior is similar to the intra-layer composition we observe in Section~\ref{subsec:ffn_pattern_aggregation}.
A possible conjecture is that the feed-forward layer acts as an elimination mechanism to ``veto'' the top prediction in the residual, and thus shifts probability mass towards one of the other candidate predictions in the head of the residual's distribution.

Finally, we manually analyze 100 random cases of last-layer composition, where the feed-forward layer modifies the residual output in the \textit{final} layer.
We find that in most cases (66 examples), the output changes to a semantically distant word (e.g.,~~~~~\nl{people} $\rightarrow$ \nl{same}) and in the rest of the cases (34 examples), the feed-forward layer's output shifts the residual prediction to a related word (e.g. \nl{later} $\rightarrow$ \nl{earlier} and \nl{gastric}~~$\rightarrow$~~\nl{stomach}).
This suggests that feed-forward layers tune the residual predictions at varying granularity, even in the last layer of the model.

\section{Related Work}
\label{sec:related_work}

Considerable attention has been given to demystifying the operation of neural NLP models. 
An extensive line of work targeted neuron functionality in general, extracting the properties that neurons and subsets of neurons capture \cite{durrani2020analyzing, dalvi2019one, rethmeier2020tx, mu2020compositional, vig2020investigating}, regardless of the model architecture or neurons' position in it. \citet{jacovi2018understanding} analyzed CNN architectures in text classification and showed that they extract key n-grams from the inputs. 

The study of the transformer architecture has focused on the role and function of self-attention layers \cite{voita2019analyzing, clark2019what, vig2019analyzing} and on inter-layer differences (i.e. lower vs. upper layers)~\cite{tenney2019bert, jawahar2019bert}. Previous work also highlighted the importance of feed-forward layers in transformers \cite{press2020improving, pulugundla2021attention, xu2020transformer}. Still, to date, the role of feed-forward layers remains under-explored.

Also related are interpretability methods that explain predictions \cite{han2020explaining, wiegreffe2019attention}, however, our focus is entirely different: we do not interpret individual predictions, but aim to understand the mechanism of transformers.

Characterizing the functionality of memory cells based on examples that trigger maximal activations has been used previously in NLP \cite{rethmeier2020tx} and vision \cite{erhan2009visualizing}.

\section{Discussion and Conclusion}
\label{sec:conclusion}

Understanding how and why transformers work is crucial to many aspects of modern NLP, including model interpretability, data security, and development of better models. Feed-forward layers account for most of a transformer's parameters, yet little is known about their function in the network. 

In this work, we propose that feed-forward layers emulate key-value memories, and provide a set of experiments showing that: (a) keys are correlated with human-interpretable input patterns; (b) values, mostly in the model's upper layers, induce distributions over the output vocabulary that correlate with the next-token distribution of patterns in the corresponding key; and (c) the model's output is formed via an aggregation of these distributions, whereby they are first composed to form individual layer outputs, which are then refined throughout the model's layers using residual connections.

Our findings open important research directions:
\begin{itemize}[leftmargin=*,topsep=3pt,itemsep=3pt,parsep=3pt]
    \item \textbf{Layer embedding space.} 
We observe a correlation between value distributions over the output vocabulary and key patterns, that increases from lower to upper layers (Section~\ref{sec:w2_as_values}).
Is this because the layer's output space transforms across layers? If so, how? 
We note that this possible transformation cannot be explained solely by the function of feed-forward layers: if the model only did a series of key-value look-ups and value-distribution aggregation via weighted addition, then a single, unifying embedding space would appear more natural. Thus, the transformation might have to do with the interplay between feed-forward layers and self-attention layers.

    \item \textbf{Beyond language modeling.} 
Our formulation of feed-forward networks as key-value memories generalizes to any transformer model, e.g. BERT encoders and neural translation models. 
We thus expect our qualitative empirical observations to hold across diverse settings, and leave verification of this for future work.

    \item \textbf{Practical implications.} A better understanding of feed-forward layers has many implications in NLP. For example, future studies may offer interpretability methods by automating the pattern-identification process; memory cells might affect training-data privacy as they could facilitate white-box membership inference \cite{nasr2019comprehensive}; and studying cases where a correct pattern is identified but then suppressed during aggregation may guide architectural novelties.
    

\end{itemize}


\vspace{2mm}
Thus, by illuminating the role of feed-forward layers, we move towards a better understanding of the inner workings of transformers, and open new research threads on modern NLP models.

\section*{Acknowledgements}

We thank Shimi Salant and Tal Schuster for helpful feedback.
This work was supported in part by the Yandex Initiative for Machine Learning, the Blavatnik Interdisciplinary Cyber Research Center (ICRC), the Alon Scholarship, and Intel Corporation.
Roei Schuster is a member of the Check Point Institute of Information Technology.
This work was completed in partial fulfillment for the Ph.D degree of Mor Geva.

\bibliographystyle{acl_natbib}
\bibliography{all}

\begin{thebibliography}{27}
\expandafter\ifx\csname natexlab\endcsname\relax\def\natexlab#1{#1}\fi

\bibitem[{Baevski and Auli(2019)}]{baevski2018adaptive}
Alexei Baevski and Michael Auli. 2019.
\newblock \href {https://openreview.net/forum?id=ByxZX20qFQ} {Adaptive input
  representations for neural language modeling}.
\newblock In \emph{International Conference on Learning Representations
  (ICLR)}.

\bibitem[{Brown et~al.(2020)Brown, Mann, Ryder, Subbiah, Kaplan, Dhariwal,
  Neelakantan, Shyam, Sastry, Askell, Agarwal, Herbert-Voss, Krueger, Henighan,
  Child, Ramesh, Ziegler, Wu, Winter, Hesse, Chen, Sigler, Litwin, Gray, Chess,
  Clark, Berner, McCandlish, Radford, Sutskever, and
  Amodei}]{brown2020language}
Tom~B Brown, Benjamin Mann, Nick Ryder, Melanie Subbiah, Jared Kaplan, Prafulla
  Dhariwal, Arvind Neelakantan, Pranav Shyam, Girish Sastry, Amanda Askell,
  Sandhini Agarwal, Ariel Herbert-Voss, Gretchen Krueger, Tom Henighan, Rewon
  Child, Aditya Ramesh, Daniel~M Ziegler, Jeffrey Wu, Clemens Winter,
  Christopher Hesse, Mark Chen, Eric Sigler, Mateusz Litwin, Scott Gray,
  Benjamin Chess, Jack Clark, Christopher Berner, Sam McCandlish, Alec Radford,
  Ilya Sutskever, and Dario Amodei. 2020.
\newblock Language models are few-shot learners.
\newblock In \emph{Proceedings of Neural Information Processing Systems
  (NeurIPS)}.

\bibitem[{Clark et~al.(2019)Clark, Khandelwal, Levy, and
  Manning}]{clark2019what}
Kevin Clark, Urvashi Khandelwal, Omer Levy, and Christopher~D. Manning. 2019.
\newblock What does {BERT} look at? {A}n analysis of {BERT}'s attention.
\newblock In \emph{BlackBoxNLP Workshop at ACL}.

\bibitem[{Dalvi et~al.(2019)Dalvi, Durrani, Sajjad, Belinkov, Bau, and
  Glass}]{dalvi2019one}
Fahim Dalvi, Nadir Durrani, Hassan Sajjad, Yonatan Belinkov, Anthony Bau, and
  James Glass. 2019.
\newblock What is one grain of sand in the desert? analyzing individual neurons
  in deep nlp models.
\newblock In \emph{Proceedings of the AAAI Conference on Artificial
  Intelligence}, volume~33, pages 6309--6317.

\bibitem[{Devlin et~al.(2019)Devlin, Chang, Lee, and
  Toutanova}]{devlin2018bert}
Jacob Devlin, Ming-Wei Chang, Kenton Lee, and Kristina Toutanova. 2019.
\newblock \href {https://doi.org/10.18653/v1/N19-1423} {{BERT}: Pre-training of
  deep bidirectional transformers for language understanding}.
\newblock In \emph{North American Association for Computational Linguistics
  (NAACL)}, pages 4171--4186, Minneapolis, Minnesota.

\bibitem[{Durrani et~al.(2020)Durrani, Sajjad, Dalvi, and
  Belinkov}]{durrani2020analyzing}
Nadir Durrani, Hassan Sajjad, Fahim Dalvi, and Yonatan Belinkov. 2020.
\newblock Analyzing individual neurons in pre-trained language models.
\newblock In \emph{Proceedings of the 2020 Conference on Empirical Methods in
  Natural Language Processing (EMNLP)}.

\bibitem[{Erhan et~al.(2009)Erhan, Bengio, Courville, and
  Vincent}]{erhan2009visualizing}
Dumitru Erhan, Yoshua Bengio, Aaron Courville, and Pascal Vincent. 2009.
\newblock Visualizing higher-layer features of a deep network.
\newblock \emph{University of Montreal}, 1341(3):1.

\bibitem[{Han et~al.(2020)Han, Wallace, and Tsvetkov}]{han2020explaining}
Xiaochuang Han, Byron~C. Wallace, and Yulia Tsvetkov. 2020.
\newblock \href {https://doi.org/10.18653/v1/2020.acl-main.492} {Explaining
  black box predictions and unveiling data artifacts through influence
  functions}.
\newblock In \emph{Proceedings of the 58th Annual Meeting of the Association
  for Computational Linguistics}, pages 5553--5563, Online. Association for
  Computational Linguistics.

\bibitem[{Jacovi et~al.(2018)Jacovi, Sar~Shalom, and
  Goldberg}]{jacovi2018understanding}
Alon Jacovi, Oren Sar~Shalom, and Yoav Goldberg. 2018.
\newblock \href {https://doi.org/10.18653/v1/W18-5408} {Understanding
  convolutional neural networks for text classification}.
\newblock In \emph{Proceedings of the 2018 {EMNLP} Workshop {B}lackbox{NLP}:
  Analyzing and Interpreting Neural Networks for {NLP}}, pages 56--65,
  Brussels, Belgium. Association for Computational Linguistics.

\bibitem[{Jawahar et~al.(2019)Jawahar, Sagot, and Seddah}]{jawahar2019bert}
Ganesh Jawahar, Beno{\^\i}t Sagot, and Djam{\'e} Seddah. 2019.
\newblock \href {https://doi.org/10.18653/v1/P19-1356} {What does {BERT} learn
  about the structure of language?}
\newblock In \emph{Proceedings of the 57th Annual Meeting of the Association
  for Computational Linguistics}, pages 3651--3657, Florence, Italy.
  Association for Computational Linguistics.

\bibitem[{Liu et~al.(2019)Liu, Gardner, Belinkov, Peters, and
  Smith}]{liu2019linguistic}
Nelson~F. Liu, Matt Gardner, Yonatan Belinkov, Matthew~E. Peters, and Noah~A.
  Smith. 2019.
\newblock \href {https://doi.org/10.18653/v1/N19-1112} {Linguistic knowledge
  and transferability of contextual representations}.
\newblock In \emph{Proceedings of the 2019 Conference of the North {A}merican
  Chapter of the Association for Computational Linguistics: Human Language
  Technologies, Volume 1 (Long and Short Papers)}, pages 1073--1094,
  Minneapolis, Minnesota. Association for Computational Linguistics.

\bibitem[{Merity et~al.(2017)Merity, Xiong, Bradbury, and
  Socher}]{merity2017pointer}
Stephen Merity, Caiming Xiong, James Bradbury, and Richard Socher. 2017.
\newblock Pointer sentinel mixture models.
\newblock \emph{International Conference on Learning Representations (ICLR)}.

\bibitem[{Mu and Andreas(2020)}]{mu2020compositional}
Jesse Mu and Jacob Andreas. 2020.
\newblock Compositional explanations of neurons.
\newblock In \emph{Proceedings of Neural Information Processing Systems
  (NeurIPS)}.

\bibitem[{Nasr et~al.(2019)Nasr, Shokri, and
  Houmansadr}]{nasr2019comprehensive}
Milad Nasr, Reza Shokri, and Amir Houmansadr. 2019.
\newblock \href {https://doi.org/10.1109/SP.2019.00065} {Comprehensive privacy
  analysis of deep learning: Passive and active white-box inference attacks
  against centralized and federated learning}.
\newblock In \emph{2019 IEEE Symposium on Security and Privacy (SP)}, pages
  739--753.

\bibitem[{Peters et~al.(2018)Peters, Neumann, Iyyer, Gardner, Clark, Lee, and
  Zettlemoyer}]{peters2018elmo}
Matthew Peters, Mark Neumann, Mohit Iyyer, Matt Gardner, Christopher Clark,
  Kenton Lee, and Luke Zettlemoyer. 2018.
\newblock \href {https://www.aclweb.org/anthology/N18-1202} {Deep
  contextualized word representations}.
\newblock In \emph{North {A}merican Chapter of the Association for
  Computational Linguistics (NAACL)}.

\bibitem[{Press et~al.(2020)Press, Smith, and Levy}]{press2020improving}
Ofir Press, Noah~A. Smith, and Omer Levy. 2020.
\newblock \href {https://doi.org/10.18653/v1/2020.acl-main.270} {Improving
  transformer models by reordering their sublayers}.
\newblock In \emph{Proceedings of the 58th Annual Meeting of the Association
  for Computational Linguistics}, pages 2996--3005, Online. Association for
  Computational Linguistics.

\bibitem[{Pulugundla et~al.(2021)Pulugundla, Gao, King, Keskin, Mallidi, Wu,
  Droppo, and Maas}]{pulugundla2021attention}
Bhargav Pulugundla, Yang Gao, Brian King, Gokce Keskin, Harish Mallidi, Minhua
  Wu, Jasha Droppo, and Roland Maas. 2021.
\newblock Attention-based neural beamforming layers for multi-channel speech
  recognition.
\newblock \emph{arXiv preprint arXiv:2105.05920}.

\bibitem[{Rethmeier et~al.(2020)Rethmeier, Saxena, and
  Augenstein}]{rethmeier2020tx}
Nils Rethmeier, Vageesh~Kumar Saxena, and Isabelle Augenstein. 2020.
\newblock Tx-ray: Quantifying and explaining model-knowledge transfer in (un-)
  supervised nlp.
\newblock In \emph{Conference on Uncertainty in Artificial Intelligence}, pages
  440--449. PMLR.

\bibitem[{Sukhbaatar et~al.(2015)Sukhbaatar, Weston, and
  Fergus}]{sukhbaatar2015end}
S.~Sukhbaatar, J.~Weston, and R.~Fergus. 2015.
\newblock End-to-end memory networks.
\newblock In \emph{Advances in Neural Information Processing Systems (NIPS)}.

\bibitem[{Sukhbaatar et~al.(2019)Sukhbaatar, Grave, Lample, Jegou, and
  Joulin}]{sukhbaatar2019}
Sainbayar Sukhbaatar, Edouard Grave, Guillaume Lample, Herve Jegou, and Armand
  Joulin. 2019.
\newblock Augmenting self-attention with persistent memory.
\newblock \emph{arXiv preprint arXiv:1907.01470}.

\bibitem[{Tenney et~al.(2019)Tenney, Das, and Pavlick}]{tenney2019bert}
Ian Tenney, Dipanjan Das, and Ellie Pavlick. 2019.
\newblock \href {https://doi.org/10.18653/v1/P19-1452} {{BERT} rediscovers the
  classical {NLP} pipeline}.
\newblock In \emph{Proceedings of the 57th Annual Meeting of the Association
  for Computational Linguistics}, pages 4593--4601, Florence, Italy.
  Association for Computational Linguistics.

\bibitem[{Vaswani et~al.(2017)Vaswani, Shazeer, Parmar, Uszkoreit, Jones,
  Gomez, Kaiser, and Polosukhin}]{vaswani2017attention}
Ashish Vaswani, Noam Shazeer, Niki Parmar, Jakob Uszkoreit, Llion Jones,
  Aidan~N Gomez, {\L}ukasz Kaiser, and Illia Polosukhin. 2017.
\newblock Attention is all you need.
\newblock In \emph{Advances in Neural Information Processing Systems (NIPS)},
  pages 5998--6008.

\bibitem[{Vig and Belinkov(2019)}]{vig2019analyzing}
Jesse Vig and Yonatan Belinkov. 2019.
\newblock \href {https://doi.org/10.18653/v1/W19-4808} {Analyzing the structure
  of attention in a transformer language model}.
\newblock In \emph{Proceedings of the 2019 ACL Workshop BlackboxNLP: Analyzing
  and Interpreting Neural Networks for NLP}, pages 63--76, Florence, Italy.
  Association for Computational Linguistics.

\bibitem[{Vig et~al.(2020)Vig, Gehrmann, Belinkov, Qian, Nevo, Singer, and
  Shieber}]{vig2020investigating}
Jesse Vig, Sebastian Gehrmann, Yonatan Belinkov, Sharon Qian, Daniel Nevo,
  Yaron Singer, and Stuart Shieber. 2020.
\newblock Investigating gender bias in language models using causal mediation
  analysis.
\newblock \emph{Advances in Neural Information Processing Systems}, 33.

\bibitem[{Voita et~al.(2019)Voita, Talbot, Moiseev, Sennrich, and
  Titov}]{voita2019analyzing}
Elena Voita, David Talbot, Fedor Moiseev, Rico Sennrich, and Ivan Titov. 2019.
\newblock \href {https://www.aclweb.org/anthology/P19-1580} {Analyzing
  multi-head self-attention: Specialized heads do the heavy lifting, the rest
  can be pruned}.
\newblock In \emph{Proceedings of the 57th Annual Meeting of the Association
  for Computational Linguistics}.

\bibitem[{Wiegreffe and Pinter(2019)}]{wiegreffe2019attention}
Sarah Wiegreffe and Yuval Pinter. 2019.
\newblock \href {https://doi.org/10.18653/v1/D19-1002} {Attention is not not
  explanation}.
\newblock In \emph{Proceedings of the 2019 Conference on Empirical Methods in
  Natural Language Processing and the 9th International Joint Conference on
  Natural Language Processing (EMNLP-IJCNLP)}, pages 11--20, Hong Kong, China.
  Association for Computational Linguistics.

\bibitem[{Xu et~al.(2020)Xu, Liu, Xiong, and van Genabith}]{xu2020transformer}
Hongfei Xu, Qiuhui Liu, Deyi Xiong, and Josef van Genabith. 2020.
\newblock Transformer with depth-wise lstm.
\newblock \emph{arXiv preprint arXiv:2007.06257}.

\end{thebibliography}

\newpage

\appendix

\section{Pattern Analysis}
\label{subsec:pattern_analysis_full_example}
Table~\ref{table:pattern_analysis_full_example} provides a fully-annotated example of 25 prefixes from the memory cell $\kk_{895}^{5}$.

\begin{table*}[t!]
    \centering
    \footnotesize
    \begin{tabular}{|c|l|}
        \hline
        1 & It requires players to press \\ \hline
        1 & The video begins at a press \\ \hline
        1 & The first player would press \\ \hline
        1 & Ivy, disguised as her former self, interrupts a Wayne Enterprises press \\ \hline
        1 & The video then cuts back to the press \\ \hline
        1 & The player is able to press \\ \hline
         & Leto switched \\ \hline
        1 & In the Nintendo DS version, the player can choose to press \\ \hline
        1 & In-house engineer Nick Robbins said Shields made it clear from the outset that he (Robbins) ``was just there to press \\ \hline
        1 & She decides not to press \\ \hline
        1 & she decides not to press \\ \hline
        1 & Originally Watson signaled electronically, but show staff requested that it press \\ \hline
        1 & At post-game press \\ \hline
        1 & In the buildup to the game, the press \\ \hline
        2 & Hard to go back to the game after that news \\ \hline
        1 & In post-trailer interviews, Bungie staff members told gaming press \\ \hline
         & Space Gun was well received by the video game \\ \hline
        1 & As Bong Load struggled to press \\ \hline
         & At Michigan, Clancy started as a quarterback, switched \\ \hline
        1 & Crush used his size advantage to perform a Gorilla press \\ \hline
        1,2 & Groening told the press \\ \hline
        1 & Creative director Gregoire \texttt{<unk>} argued that existing dance games were merely instructing players to press \\ \hline
        1,2 & Mattingly would be named most outstanding player that year by the press \\ \hline
        1 & At the post-match press \\ \hline
        1,2 & The company receives bad press  \\ \hline
    \end{tabular}
    \vspace*{1mm}
    \newline
    \begin{tabular}{|c|l|l|}
        \hline
        \textbf{ID} & \textbf{Description} & \textbf{shallow / semantic} \\ \hline
        1 & Ends with the word ``press''  & shallow \\ \hline
        2 & Press/news related & semantic \\ \hline
    \end{tabular}
    \caption{A pattern annotation of trigger examples for the cell memory $\kk_{895}^{5}$. Trigger examples are annotated with repetitive patterns (upper table), which are classified as ``shallow'' or ``semantic'' (bottom table).}
    \label{table:pattern_analysis_full_example}
\end{table*}

\section{Implementation details}
\label{subsec:implementation_details}
In this section, we provide further implementation details for reproducibility of our experiments.

For all our experiments, we used the language model of \citet{baevski2018adaptive} (247M parameters) trained on WikiText-103 \cite{merity2017pointer}. Specifically, we used the model \texttt{transformer\_lm.wiki103.adaptive} trained with the fairseq toolkit\footnote{\url{https://github.com/pytorch/fairseq}}.

WikiText-103\footnote{\url{https://blog.einstein.ai/the-wikitext-long-term-dependency-language-modeling-dataset/}} is a well known language modeling dataset and a collection of over 100M tokens extracted from Wikipedia.
We used spaCy\footnote{\url{https://spacy.io/}} to split examples into sentences (Section~\ref{sec:w1_as_keys}).

\end{document}